\documentclass[10pt,twocolumn,letterpaper]{article}

\usepackage[pagenumbers]{wacv} 

\usepackage{graphicx}
\usepackage{amsmath}
\usepackage{amssymb}
\usepackage{booktabs}
\usepackage{times}
\usepackage{epsfig}
\usepackage{graphicx}
\usepackage{amsmath}
\usepackage{amssymb}
\usepackage{tabularx,booktabs}
\usepackage{bm}

\usepackage{soul}
\usepackage{color}
\definecolor{dkgreen}{rgb}{0,0.6,0}
\definecolor{gray}{rgb}{0.5,0.5,0.5}
\definecolor{mauve}{rgb}{0.58,0,0.82}

\usepackage[pagebackref,breaklinks,colorlinks]{hyperref}
\usepackage[capitalize]{cleveref}
\crefname{section}{Sec.}{Secs.}
\Crefname{section}{Section}{Sections}
\Crefname{table}{Table}{Tables}
\crefname{table}{Tab.}{Tabs.}
\usepackage[dvipsnames,svgnames]{xcolor}
\usepackage{changes}

\def\wacvPaperID{6} 
\def\confName{WACV}
\def\confYear{2024}

\begin{document}

\title{Fingerspelling PoseNet: Enhancing Fingerspelling Translation with Pose-Based Transformer Models}

\author{Pooya Fayyazsanavi, Negar Nejatishahidin ,  and Jana Ko{\v{s}}eck{\'a}\\
George Mason University\\
{\tt\small \{pfayyazs, nnejatis, kosecka\}@gmu.edu}
}


\maketitle

\begin{abstract}
We address the task of American Sign Language fingerspelling translation using videos in the wild. 
We exploit advances in more accurate hand pose estimation and propose a novel architecture that leverages the transformer based encoder-decoder model enabling seamless contextual word translation.
The translation model is augmented by a novel loss term that accurately predicts the length of the finger-spelled word, benefiting both training and inference. We also propose a novel two-stage inference approach that re-ranks the hypotheses using the language model capabilities of the decoder.
Through extensive experiments, we demonstrate that our proposed method outperforms the state-of-the-art models 
on ChicagoFSWild and ChicagoFSWild+ achieving more than 10\% relative improvement in performance. Our findings highlight the effectiveness of our approach and its potential to advance fingerspelling recognition in sign language translation. Code is also available at \url{https://github.com/pooyafayyaz/Fingerspelling-PoseNet}.

\end{abstract}

\vspace{-7pt}
\section{Introduction}
American Sign Language (ASL) is a complex and expressive visual language, that relies on hand gestures, facial expressions, and body movements to convey meaning.  It has its own unique grammar and syntax. In comparison to the remarkable advancements achieved in Automatic Speech Recognition (ASR), sign language recognition and translation are still in its early stages of development. 
It encompasses diverse sub-tasks, including fingerspelling translation, word-level recognition, and continuous translation. Sign language translation faces challenges such as the availability of limited paired data for training models and the complexity of extracting effective representation from visual modality. 

This paper focuses on fingerspelling translation, which involves accurately detecting and interpreting the specific hand poses and movements used to spell out individual letters. According to \cite{padden2003alphabet}, fingerspelling accounts for approximately 12-35\% of communication in American Sign Language (ASL). This functionality is crucial for recognizing proper nouns, technical terms, and words that do not have dedicated signs. \\ 

\begin{figure}[t]
\centerline{\includegraphics[width=\linewidth]{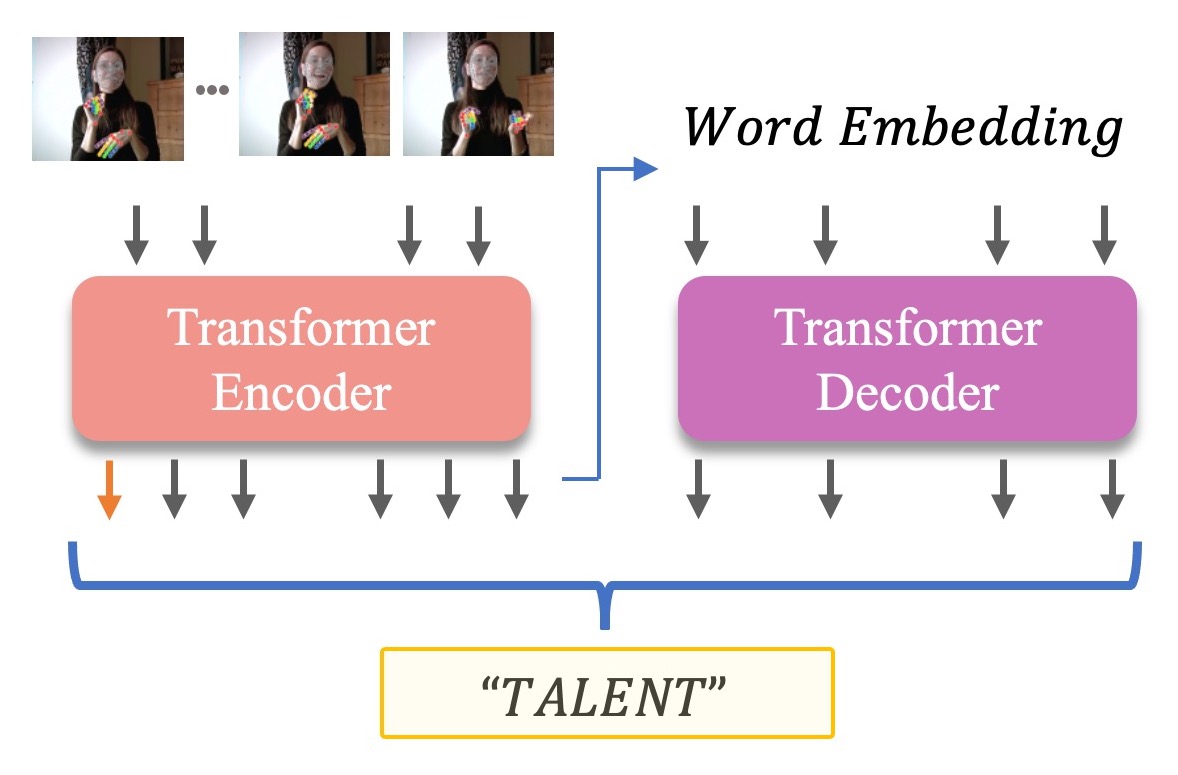}}
\caption{Our overall architecture utilizes a sequence of hand poses as the input. The encoder and decoder components work in conjunction to generate the final output ("TALENT").}
\vspace{-15pt}
\label{ff}
\end{figure}

There are some unique challenges in American fingerspelling translation. It uses a single hand which involves relatively small and quick motions of the hand and fingers, as opposed to the typically larger arm motions involved in other ASL sub-tasks. Therefore, fingerspelling can be difficult to analyze with standard approaches. 
Current fingerspelling methods \cite{fs18iccv, fingeriter, prajwal2022weaklyfinger, pannattee2021novel} primarily rely on appearance-based techniques and often face limitations due to high variability among signers, including differences in speed, hand appearance, and other motion variations before and after signing. In contrast, pose-based methods have the potential to be robust to these variations and offer data efficiency while addressing privacy concerns.\\
In this work, we propose a novel pose-based approach using encoder-decoder transformer model summarized in Figure \ref{ff}. Transformers have demonstrated remarkable success in various natural language processing tasks by effectively capturing long-range dependencies and contextual information. Using transformers in the domain of fingerspelling recognition, we were able to exploit their language modeling capabilities and achieve significant improvements on the ChicagoWild~\cite{fs18slt} and ChicagoWild+~\cite{fs18iccv} dataset. These datasets consist of a diverse range of hand gestures corresponding to individual letters, captured from multiple signers in various environments. In summary, the contributions of the proposed approach are summarized as follows:

\begin{itemize}
    \item Transformer-based architecture that combines Connectionist Temporal Classification (CTC) and language modeling for fingerspelling. The model captures contextual information and enables effective language modeling and seamless translation within a single framework.
    \item Introducing a novel loss term for predicting the word length that enhances translation accuracy and robustness, particularly in cases of missing letters. This improvement benefits both training and inference.
    \item Novel two-stage inference approach exploiting the learned language model for re-ranking the hypotheses.
    \item Our method surpasses existing SOTA models, achieving over 10\% relative improvement in finger spelling translation performance.
\end{itemize}

\begin{figure*}[tbp]
\centerline{\includegraphics[width=0.95\linewidth]{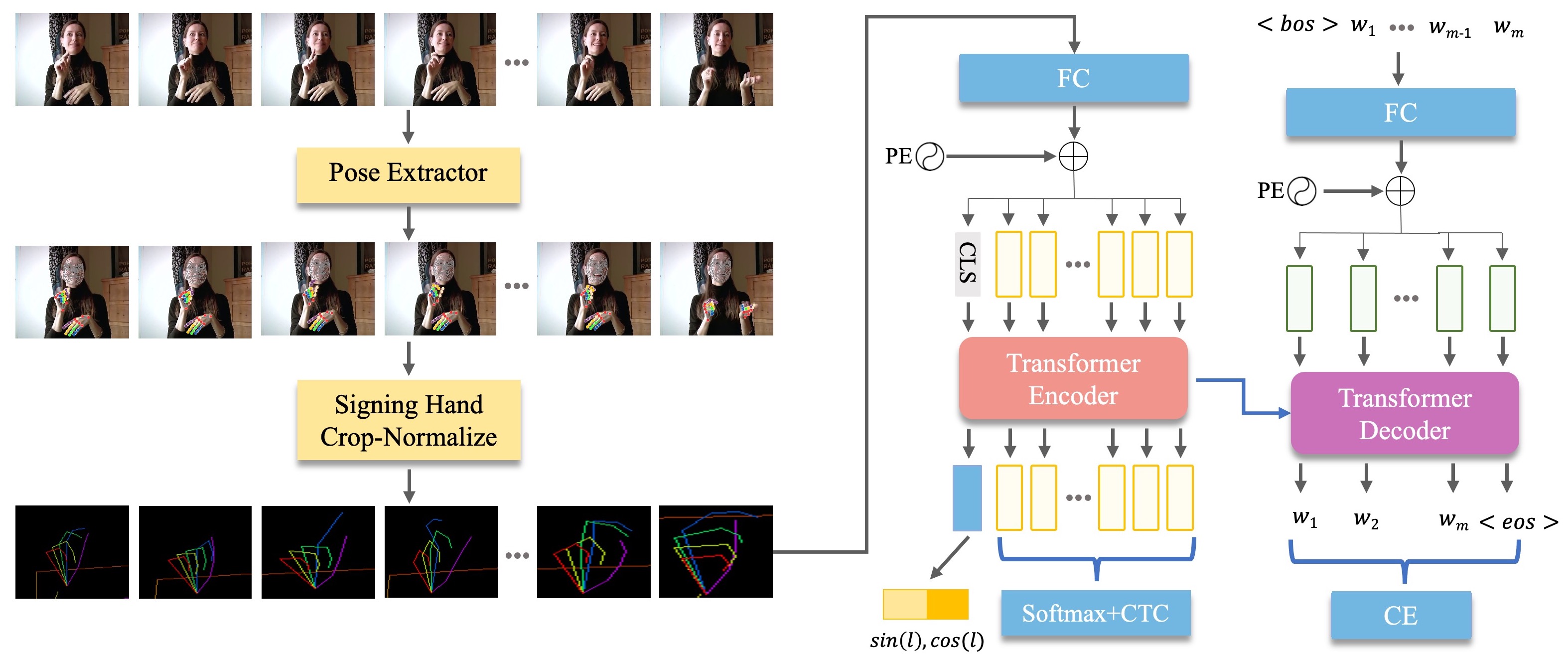}}
\caption{The figure illustrates our overall approach. Raw frames are processed to extract keypoints, which are then cropped and normalized. These normalized keypoints are passed through a fully connected layer for upscaling and fed into a Transformers encoder. Additionally, a special token predicts the letter size using normalized sine and cosine values. On the decoder side, the letter sequence is augmented with {\tt BOS} and {\tt EOS} tokens which generate the subsequent tokens in an autoregressive manner.}
\label{fig:arc}
\end{figure*}

\section{Related Work}
Early works on sign language recognition from video focused on isolated signs~\cite{10.1007/978-3-030-66096-3_18,Wadhawan2021,OZ20111204,bohavcek2022sign,li2020word,du2022full,hosain2021hand}, where individual signs, words or letters, are recognized in isolation. More recent advancements in the field have shifted towards continuous sign language recognition~\cite{duarte2021how2sign,shi2022open,cui2017recurrent,subunet,li2022multi}, aiming to parse and translate continuous signing sequences. 
%
%
More recently deep learning techniques have been applied to fingerspelling recognition, leveraging the power of convolutional neural networks (CNNs), recurrent neural networks (RNNs)~\cite{schuster1997bidirectional,fayyazsanavi2023u2rle}, and Transformers~\cite{vaswani2017attention}. 
The choice of representation plays a crucial role in modeling sign language, as it directly impacts the performance and robustness of recognition and translation systems. In RGB-based approaches, a 2D/3D convolutional neural network backbone pre-trained on datasets like ImageNet~\cite{deng2009imagenet}, DeepHand~\cite{sinha2016deephand}, or activity recognition datasets~\cite{kay2017kinetics} is commonly employed. For continuous translation, the backbone can be pre-trained using word-level data, as demonstrated in~\cite{shi2022open, pu2019iterative}. RGB-based representations are often susceptible to lighting conditions, background clutter, and high visual domain variation. These challenges impact the accuracy of recognition and require a large amount of data for training from scratch. 

Skeleton-based representations use spatial positions of joints and body landmarks of the signer. These models ~\cite{jiang2021skeleton,bohavcek2022sign,moryossef2021evaluating,parelli2020exploiting} utilize off-the-shelf pose estimation methods and then learn spatio-temporal features on the top of these 2D or 3D keypoint coordinates. Authors in~\cite{parelli2020exploiting} proposed simple linear layers to lift the 2D 
keypoints into the 3D instead of using 3D pose estimation. 

Previous works~\cite{hosain2020finehand,santhalingam2020body,santhalingam2019sign,fs18slt,fs18iccv,subunet,cui2017recurrent} have extensively employed LSTM and RNN architectures for various sign language tasks and vary in the types of input, model architectures, and fusion strategies when combining multiple channels of information. These models excel in capturing sequential dependencies and have been widely adopted for their ability to model temporal information in video data. 
Following the advancements in Natural Language Processing (NLP), transformer-based approaches have gained significant attention in sign language processing~\cite{shi2022open,bohavcek2022sign,moryossef2021evaluating,prajwal2022weaklyfinger}. Transformers excel at capturing long-range dependencies\cite{nejatishahidin2023graph} and contextual information, making them suitable for modeling the complex dynamics of sign language. The transformers-based models either use features learned from the video frames~\cite{camgoz2020sign,du2022full} or the 2D/3D pose estimates~\cite{bohavcek2022sign,moryossef2021evaluating}. The supervision can be both on the encoder or decoder side. The decoder's auto-regressive component is effective in modeling the linguistic 
structure both in case of RNNs~\cite{camgoz2018neural} and transformer architectures~\cite{camgoz2020sign}. 
in the case of word-level classification task, the decoder of the transformer decodes the class query~\cite{bohavcek2022sign}. \\

Despite the progress in both word-level recognition and continuous translation, there remains a gap in the literature concerning the specific task of 
fingerspelling translation. Fingerspelling translation in real-world scenarios has been extensively explored in~\cite{fs18iccv, fs18slt, prajwal2022weaklyfinger,shi-etal-2022-searching}. These studies collected videos from YouTube and Deaf social media platforms to capture diverse fingerspelled words in natural contexts. In the work by~\cite{fs18slt} the authors developed a hand detection method to locate the signing hand within the video frames, followed  by training a CNN-LSTM model for translation purposes. The follow-up work~\cite{fs18iccv} presented an end-to-end approach that bypassed the explicit hand detection step and proposed an iterative attention mechanism, leveraging a 2D-CNN to extract visual features from individual frames, which were used as an input to RNN. To enhance representation learning,~\cite{pannattee2021novel} introduced a Siamese network architecture to distinguish between similar and dissimilar hand shapes. These works primarily focus on videos with exclusive fingerspelling content, which is a limitation in realistic scenarios where the exact occurrence of fingerspelling is unknown. In~\cite{shi2021fingerspelling} the model first detects segment proposals in the video, and subsequently performs recognition of these segments using the CTC \cite{graves2006connectionist} loss. In~\cite{prajwal2022weaklyfinger}, the authors employ a multi-stage training strategy to overcome the need for labeled segmentation, leveraging additional cues such as mouthing.
In \cite{gajurel2021fine} authors explore the use of optical flow as additional input to the Transformers Encoder. Meanwhile, in \cite{papadimitriou20_interspeech}, translation is approached through multi-modal fusion involving pose, optical flow, and CNN features. The work of \cite{kabade2023american} employs an attention-based CNN approach for generating spatial features, utilizing optical flow as a prior for LSTM modeling. Fingerspelling often encounters the issue of distinct letters sharing highly similar hand-shapes, leading to ambiguities.  This issue is addressed by \cite{li-etal-2023-contrastive-token} by modifying the transformers Encoder-Decoder to effectively discern these ambiguities in visual representations.

In recent work~\cite{shi2022open}, a new dataset combines continuous sign language and fingerspelling, offering rich training data. Using pre-trained networks and multi-modal transformers, the study reveals a BLEU-4 score decrease (7.74 to 6.33) in videos containing fingerspelling. This reveals a limitation in existing models and emphasizes the potential for improvement in this area.



\section{Approach}
The aim of a finger-spelling translation system is to convert a collection of video frames $I=\left\{{I}_1, {I}_2, \ldots, {I}_T\right\}$ into a letter sequence, $W =\{w_1, w_2, \ldots, w_L\}$, thus translating the entire video sequence. We have access to a set of $n$ pairs $\{I, W\}$ where $I$ is a video and $w$ is the corresponding label. Our transformer-based model uses the sequence of hand landmarks extracted from the video frames as input. Additionally, our model incorporates a novel loss function designed to predict the length of the word. The overall architecture is outlined in Figure \ref{fig:arc}. In the following sections, each of these components will be described in detail.


\subsection{Input Representation and Pre-processing}

\noindent\textbf{Pose Estimation.}
\label{sec:sigext}To estimate the human body pose from the video frames, various off-the-shelf methods can be employed. While previous works mainly used on OpenPose~\cite{openpose}, this study utilizes the Google MediaPipe Holistic framework~\cite{lugaresi2019mediapipe}. In Section \ref{ab:Pose} we present the effect of different pose estimation methods on the final translation task. 
MediaPipe provides 543 body landmarks MediaPipe (33 body joints pose landmarks, 468 face landmarks, and 21 hand landmarks per hand), where the hand joints are specifically employed for training the model. Each landmark comes with a $confidence$ value and $3$D coordinates consisting of $x$, $y$, and $z$. In this work, only the $x$ and $y$ coordinates are utilized for training purposes.\\

\noindent
\textbf{Signing Hand Detection.}
\label{sec:signdet}
In American Sign Language (ASL), finger spelling is performed using only one hand. Consequently, one of the initial steps in the pre-processing stage involves identifying which hand does the fingerspelling. Two techniques were employed to determine the hand involved in the process. First, the finger joint positions obtained and used to analyze the movements and gestures of each hand. The dominant hand typically exhibits more variability(difference between consecutive frames) in joint movements. 
\vspace{-10pt}
\begin{equation}
V=\sum_{t=1}^{T}\sum_{j=0}^J P^t_j-P^{t-1}_j
\vspace{-5pt}
\end{equation}

Here, $P^t_j$ denotes the $j$-th hand joint at frame $t$, with $T$ representing the total number of frames and $J$ representing the count of hand joints. The value $V$ is subsequently compared for the right and the left hand, and the larger value is chosen to determine the dominant hand. To further improve the accuracy of this heuristic, we leverage the consistency observed in signers' hand usage type across different videos. We check the current predictions with the previous ones made by the same signer. This approach takes advantage of the fact that signers tend to consistently use the same hand for fingerspelling in all of their videos. By considering the past patterns of hand usage for each signer, we can refine the predictions and achieve more accurate results.

\label{sec:norm}
\noindent
The estimated hand landmarks need to be normalized before the training. Normalizing pose data ensures that different poses are represented consistently across different individuals or scenarios like scale, orientation, and position.\\

\noindent
\textbf{Hand Origin.} To normalize all the $x$ and $y$ coordinates, we utilize the wrist landmark origin of the hand coordinate system and normalize other joints employing the following procedure:
$f_{\text {origin}}(x,y)= (x - x_{origin}, y - y_{origin})$
where $x_{origin}, y_{origin}$ are the $x,y$ location of wrist landmark.\\

\noindent
\textbf{Mirror.} In the case of signers utilizing the left hand, we employ a mirroring technique to adjust the hand landmarks in the following manner:
$f_{\text {mirror}}(x)= -x + \max(X)$
Let $x$ represent the x-coordinate, and $X$ denote the array containing all the $x$ coordinates in one frame.\\

\noindent
\textbf{Scaling.} In order to address the scaling issue, we uniformly resize the hand bounding box to a dimension of $1 \times 1$. This transformation guarantees that all values are scaled within the range of 0 to 1, with the maximum value set to 1 and the remaining values adjusted proportionally. By applying this transformation, we ensure consistent scaling across all hand instances as follows: 
$f_{\text {scale}}(x,y)= \left(\frac{x}{\max(X)}, \frac{y}{\max(Y)} \right)$
Where the $X,Y$ represent the array that contains all the $x$ and $y$ locations in one frame.\\

Lastly, all the hand joint coordinates are normalized by subtracting the mean and dividing by the maximum absolute value. This process ensures that the values are scaled in the range of $[-0.5, 0.5]$ while being centered around zero.

\subsection{Model Architecture}
\label{sec:method}
Our approach utilizes transformer-based encoder-decoder architecture, initially proposed in~\cite{vaswani2017attention} and depicted in Figure~\ref{fig:arc}.
The input to our system is a sequence of normalized body poses, each containing 21 keypoint coordinates. The extraction of hand poses from the video involves applying the procedure described in Section \ref{sec:sigext}, utilizing MediaPipe and subsequent pre-processing steps. The encoder takes in a tensor $P=\left\{p_1, p_2, \ldots, p_T\right\}$ of size $T \times 21 \times 2$, which is then flattened to yield a tensor of size $T \times 42$. Subsequently, a learnable positional encoding is added to the vector of poses.
The sequence then passes through the self-attention module and a feed-forward network composed of two layers, to capture contextual information within the pose sequence. The self-attention module has 8 attention heads in each of the 3 encoder layers. \\

\noindent
\textbf{Length Token.} In the transformers encoder block we incorporate a learnable parameter token and concatenate it with the vector of poses. This output token is then mapped into a vector of size 2 using a fully connected layer in the output. The role of this token is to predict the number of letters in the word in sign language fingerspelling. We observed that existing models often struggle with accurate prediction of certain letters, leading to performance limitations. By introducing this token, we aim to improve the prediction of missing letters. Furthermore, during the inference, we leverage this prediction to enhance the accuracy and robustness of our model's predictions. 
To generate the ground truth data for this prediction, we transform the length in:
{\small
\begin{equation}
len=\left[\sin \left( 2\pi *\left( \frac{L}{30}-0.5\right)\right), \cos \left( 2\pi *\left(\frac{L}{30}-0.5\right)\right)\right]
\label{eq:length}
\end{equation}
}
where $L$ represents the length of the word. Initially, we normalize the length values, with $L=30$ being the longest word, transform them between $[-\pi, \pi]$ and compute the sine and cosine of these normalized lengths. 
By using sine and cosine representations the errors in length prediction are mapped to points on a unit circle enabling more balanced treatment and making the contribution of the errors less sensitive to the absolute scale of the words.  

On the decoder side, the model takes in the sequence of letters. We first tokenize the letters, augment them with the beginning-of-sequence {\tt BOS} and end-of-sequence {\tt EOS} tokens, and add the positional embeddings to the tokens representing letters. The augmented and embedded sequence $W_{\text {word }}=\left\{w_1, w_2, \ldots, w_L\right\}$, of length $L$, is then passed through the decoder. The decoder employs a masked attention mechanism, where each token can attend to only the preceding tokens, preventing the model from accessing future information during training. This enables the decoder to generate tokens autoregressively, attending only to the already generated parts of the sequence. Following the masked attention step, the decoder further utilizes self-attention mechanisms, allowing each token to attend to all other tokens in the sequence capturing global dependencies and context. The self-attention mechanism facilitates the decoder in generating the output tokens one at a time, progressively constructing the final output sequence. The decoder has 3 layers with 8 attention heads. \\

\subsection{Loss Functions}
In our fingerspelling translation task using a transformer encoder-decoder, we employed three distinct loss functions 
that will be discussed in detail next. \\

\noindent
\textbf{CTC Loss.} On the encoder side, where the input comprises a sequence of hand poses without explicit alignments between the poses and the target sign language letters. We use Connectionist Temporal Classification (CTC) loss function. The CTC loss models all possible alignments between the hand shapes and the sign language letters without requiring explicit alignment supervision.
\begin{equation}
\mathcal L_{\mathrm{CTC}}=-\log p(W \mid P)
\end{equation}
where $P$ is the vector of poses and $W$ is the target sequence of labels. In more detail:
\begin{equation}
\mathcal L_{C T C}=-\log \sum_{A \in \mathcal{A}_{P, W}} \prod_{t=1}^T p\left(c_t \mid P\right)
\end{equation}
where, $A \in \mathcal{A}_{P, W}$ denotes the set of valid alignments corresponding to the target sequence $W$, and $p\left(c_t \mid P\right)$ denotes the probability of corresponding letter at timestep $t$ of the input sequence. The term $p\left(c_t \mid P\right)$ is the output of the encoder at each timestep, where $c_t$ is the probability of the letter at the output of the softmax layer.
\newline

\noindent\textbf{MSE Loss.} To further enhance the performance and learning capabilities of the model, we introduced a learnable parameter to predict the length of the letters during training. This additional parameter allowed the model to gain a better understanding of the variations in letter sizes within sign language. By training this parameter using a Mean Squared Error (MSE) loss function, the model could improve its ability to accurately predict the length of the letters. The length prediction could also be leveraged during the inference stage, aiding in generating more accurate and visually consistent translations.

\begin{equation}
\mathcal L_{\mathrm{MSE}}=\frac{1}{N} \sum_{i=1}^N \frac{1}{2} \sum_{j=1}^2\left(\widehat{\operatorname{len}^i_{ j}}-\operatorname{len}^i_{j}\right)^2
\end{equation}
In this equation, $[\widehat{len}^i_1, \widehat{len}^i_2]$ represents the predicted word length, which is a vector of size $2$ (sine and cosine) (see Eq. \ref{eq:length}). The ground truth length of $i$-th example is denoted as $[{len}^i_1, len^i_2]$ and $N$ represents the batch size.
\newline

\noindent
\textbf{Cross Entropy.} On the decoder side, the task involved generating the sign language letter translation one letter at a time, following an auto-regressive approach. To optimize the decoder's performance in this auto-regressive task, we utilized a cross-entropy loss function. The cross-entropy loss encouraged the model to produce more accurate and contextually appropriate letter predictions.

\begin{equation}
\mathcal L_{CE}=-\frac{1}{M} \sum_{i=1}^M y_i \cdot \log \left(\hat{y}_i\right)
\end{equation}
where $y_i$ is the ground truth label and $\hat{y_i}$ is the softmax probability for the $i^{th}$ class, $M$ represents the total number of classes.\\
The total loss is calculated as:

\begin{equation}
\label{eq:loss}
\mathcal{L}=\lambda \mathcal{L}_{\text {CTC }}+\mathcal{L}_{\text {CE }}+ \mathcal{L}_{\text {MSE }}
\end{equation}
where $\lambda$ is utilized to regulate the relative contributions of loss components.

\noindent
\subsection{Re-ranking Inference}
\label{rerankinfer}
Throughout the prediction process, the encoder employs a greedy decoding strategy to generate the likelihood of each letter for every frame.
\begin{equation}
\hat{W}=\underset{{\mathbf{W}} \in \boldsymbol{w}^*}{\operatorname{argmax}} \prod_{t=1}^T p_{ctc}\left(c_t \mid \varepsilon(P)\right)
\end{equation}

The beam search then refines the generated candidates by considering their likelihood and selecting the most probable sequences, We refer to this set of $k$ predictions as our hypotheses. In the proposed model, the contextualized features $\varepsilon(P)$ are obtained from the encoder, where $P$ represents the input pose sequence. The sequence length is denoted by $T$. At each timestep $t$, $a_t$ represents the probability of characters. However, utilizing beam search solely on the encoder side fails to capitalize on the potential of a language model. The language model captures the probability distribution of letters based on the generated letters up to a given point.

\begin{equation}
p\left(w_1, w_2, \ldots, w_L\right)=\prod_{i=1}^L p\left(w_i \mid w_1, w_2, \ldots, w_{i-1}\right)
\end{equation}

 Therefore we employ autoregressive decoding on the decoder side. With this approach, the model generates the output sequence token by token, taking into account the previously generated tokens. This autoregressive process enables the model to capture the context and dependencies within the sequence, leading to coherent and contextually appropriate predictions.

\begin{equation}
\hat{w_t}=\mathcal{D}\left(\hat{w}_{1: t-1}, \varepsilon(P)\right)
\end{equation}

Using the input pose vector $P$, the encoder ($\varepsilon$) generates contextualized tokens. The decoder starts with the {\tt SOS} token and proceeds to generate subsequent tokens until either the model generates the {\tt EOS} token or the maximum length is reached.

However, there are some drawbacks of using this method. Firstly, generating a meaningful sequence, especially in the case of fingerspelling with limited available data, necessitates a substantial amount of training data. Secondly, prior research \cite{ren2020study, he2019hard, song2021non}
has shown that the decoder is more sensitive to target-side information rather than source-side information. Consequently, even a minor mis-recognition can significantly degrade the overall predicted performance. On the other hand, employing a separate language model, such as \cite{fs18slt}, might overlook the rich contextual information encoded by the encoder and focus solely on language aspects.\\

To address these limitations, we propose a hybrid approach that combines the strengths of both methods. During the decoding process, we utilize the CTC with beam decoding technique to generate a set of hypotheses. However, rather than relying solely on an autoregressive method, we take these hypotheses as input and employ a re-ranking strategy based on the generated probabilities on the decoder. This integration allows us to benefit from the contextualized encoder features while leveraging the hypothesis generation capability of the CTC.\\

Furthermore, we incorporate the predicted length to improve the ranking process. This aspect proves particularly beneficial, as one of the limitations is the potential omission of certain letters. By integrating the predicted length, our model generates more consistent predictions and improves the overall performance.
\begin{equation}
\small
\hat{W}=\underset{W \in w^*}{\operatorname{argmax}} \log p_{\mathrm{ctc}}(W \mid P)+\beta \log p_{\operatorname{lm}}(W \mid \varepsilon(P))-\gamma E_L
\label{infereq}
\end{equation}
Where,
\begin{equation}
\small
E_L = \left|\hat{L}-L_Y\right|
\end{equation}

In our approach, the decoder, denoted as $p_{\text{lm}}(W \mid \varepsilon(P))$, takes as input the generated hypotheses from the beam search. $\hat{L}$ is the predicted Length token and $L_Y$ is length of the hypotheses generated on the encoder.

\begin{table}[t]
\centering
\begin{tabular}{c|c}
\hline Decoding Strategy  & Letter Accuracy\%  \\
\hline \multicolumn{2}{c}{ Encoder only } \\
\hline Encoder Only(CTC) Greedy  & $57.3$  \\
\hline Encoder Only(CTC) + Beam   & $58.5$   \\
\hline Encoder Only(CTC) + LSTM \cite{fs18slt}  & $59.8$  \\

\hline \multicolumn{2}{c}{ Encoder-Decoder } \\
\hline Encoder-Decoder(only CE) & $54.6$  \\
\hline Encoder-Decoder(CTC + CE) & $56.3$  \\
\hline Ours  & \bm{$66.3$} \\
\hline
\end{tabular}
\caption{Comparison of Training and Decoding Strategies for FingerSpelling translation. For training, we can incorporate CTC loss, CE, or both. During inference, decoding includes auto-regressive on the decoder or beam search decoding on the encoder side.}
\label{tab:dec}
\end{table}%

\section{Experiments}
We report the results of our approach on ChicagoFSWild\cite{fs18slt}, and ChicagoFSWild+\cite{fs18iccv} datasets. We also provide information regarding the training schema, inference ranking, used datasets, and our ablation study. 

\subsection{Training}
We implement our model using PyTorch \cite{paszke2019pytorch} framework. The Adam optimizer \cite{kingma2014adam} is employed to train our network with $\beta_1 = 0.9, \beta_2 =0.999$. The network is trained on one NVIDIA GeForce GPU for $20$ epochs on both ChicagoFSWild and ChicagoFSWild+ datasets, with a batch size of $1$. In addition, we set hyper-parameters in Eq. \ref{eq:loss} as $\lambda= 5$. All the hyperparameters are determined using the validation set.

\subsection{Dataset}
The Chicago Fingerspelling Dataset\cite{fs18slt} is a collection of videos that feature individuals performing American Sign Language (ASL) fingerspelling. This dataset was created “in the wild”, using videos collected from websites. ChicagoFSWild includes $7304$ ASL sequences by 160 signers, while ChicagoFSWild+\cite{fs18iccv} contains $55,232$ sequences by 260 signers. The datasets offer video-level annotations but lack individual frame-level segmentation.

\subsection{Inference}
The inference stage plays a crucial role in generating accurate predictions. In this section, we present three main stages employed during the inference stage, namely CTC with beam search, autoregressive decoding, and our re-ranking inference. Following the prior works \cite{shi-etal-2022-searching,fs18iccv,fs18slt,li-etal-2023-contrastive-token} we evaluate the performance
based on the metrics of letter accuracy $\text { ErrorRate }=\frac{(S+D+I)}{N}$, where
$S, D, I$ are the number of substitutions, deletions, and insertions in the alignments, and N is the number of letters.\\

\noindent
\textbf{CTC with Beam Search.}
First, the CTC with beam search technique is commonly used to generate multiple hypotheses or candidate sequences. Table \ref{tab:dec} presents the results for non-autoregressive decoding using the CTC approach. Our experiments conducted in two scenarios. In the first scenario, we performed greedy decoding, selecting the most probable character at each time step. Secondly, in order to enhance the prediction quality, we incorporate beam search with a beam width of $5$ to consider multiple hypotheses, as demonstrated in Table \ref{tab:dec}. \\

\noindent
\textbf{CTC with Language Model.}
In this experiment, we leverage the language model trained specifically for finger spelling, as introduced in \cite{fs18slt}. This dedicated language model is employed to refine the generated hypotheses, leading to improved results, as demonstrated in Table \ref{tab:dec}. The language model consists of an LSTM trained separately on the training set of labels. \\

\noindent
\textbf{Autoregressive Decoding.}
Another approach is to employ autoregressive decoding on the decoder side. With this approach, we solely rely on the decoder to generate the output sequence. The utilization of the CTC loss during training leads to improved results during inference, as demonstrated in Table \ref{tab:dec}.
However, as shown in Table \ref{tab:dec}, the performance of the models in this scenario still lags behind that of the non-autoregressive counterparts. This outcome was expected, as explained in Section \ref{rerankinfer}.\\

\noindent
\textbf{Our Method.}
In our approach, we aim to leverage the strengths of language models while giving importance to the contextualized features from the encoder as described in Section \ref{rerankinfer}. The values of $\beta$ and $\gamma$ are assigned as $0.4$ and $1.2$, respectively in Eq. \ref{infereq}. As shown in Table \ref{tab:dec}, the model can outperform all other inference strategies.

\begin{table}[h]
\centering
\begin{tabular}{lcc}
\hline \textbf{Model} &  \footnotesize \textbf{FSWild}~\cite{fs18slt} & \footnotesize \textbf{FSWild+}~\cite{fs18iccv} \\
\hline \small Resnet Whole Frame & $22.3 \%$ & $24.7 \%$ \\
\small Hand Det.+ CNN + RNN~\cite{fs18slt}  & $41.9 \%$ & $41.2 \%$ \\
\small Iterative Attention + LM~\cite{fs18iccv} & $45.1 \%$ & $46.7 \%$ \\
\small Weakly Supervised~\cite{pannattee2021novel} & $48 \%$ & - \\
\small Fine-Grained Attention~\cite{gajurel2021fine} & $48.36 \%$ & - \\
\small TDC-SL~\cite{papadimitriou20_interspeech} & $50 \%$ & - \\
\small Attention(optical flow+Res)~\cite{kabade2023american} & $57.84 \%$ & - \\
\small FSS-Net~\cite{shi-etal-2022-searching} & $52.5 \%$ & $64.4 \%$ \\
\small CtoML~\cite{li-etal-2023-contrastive-token} & $54.9 \%$ & - \\

\small Ours(Enc-Dec Transformers) & \bm{$66.3 \%$} & \bm{$71.1 \%$} \\
\hline
\end{tabular}
\caption{Comparing different models on the test set of the ChicagoWild \cite{fs18slt} and the ChicagoWild+ \cite{fs18iccv} datasets, we evaluate the performance using the metric of Letter Accuracy ($\%\uparrow$).}
\label{tab:fullresult}
\end{table}

\subsection{Result}
In this section, we present the results of our experiments and evaluations conducted to assess the performance of our proposed method. We aim to provide an analysis and interpretation of the outcomes obtained, showcasing the advancements and contributions made toward the problem. We adopted the train/val/test split introduced in the original paper\cite{fs18slt}. The results, as shown in Table \ref{tab:fullresult}, are compared with various models on both datasets, demonstrate that our model surpasses all other models by a significant margin.
Our study also investigates the impact of various factors, including model architectures, different inference techniques, and hyperparameters to establish a robust and reliable framework for tackling the challenges at hand. All the ablations are using the Chicago Wild~\cite{fs18slt} dataset. 

\begin{figure*}[tbp]
\centerline{\includegraphics[width=\linewidth]{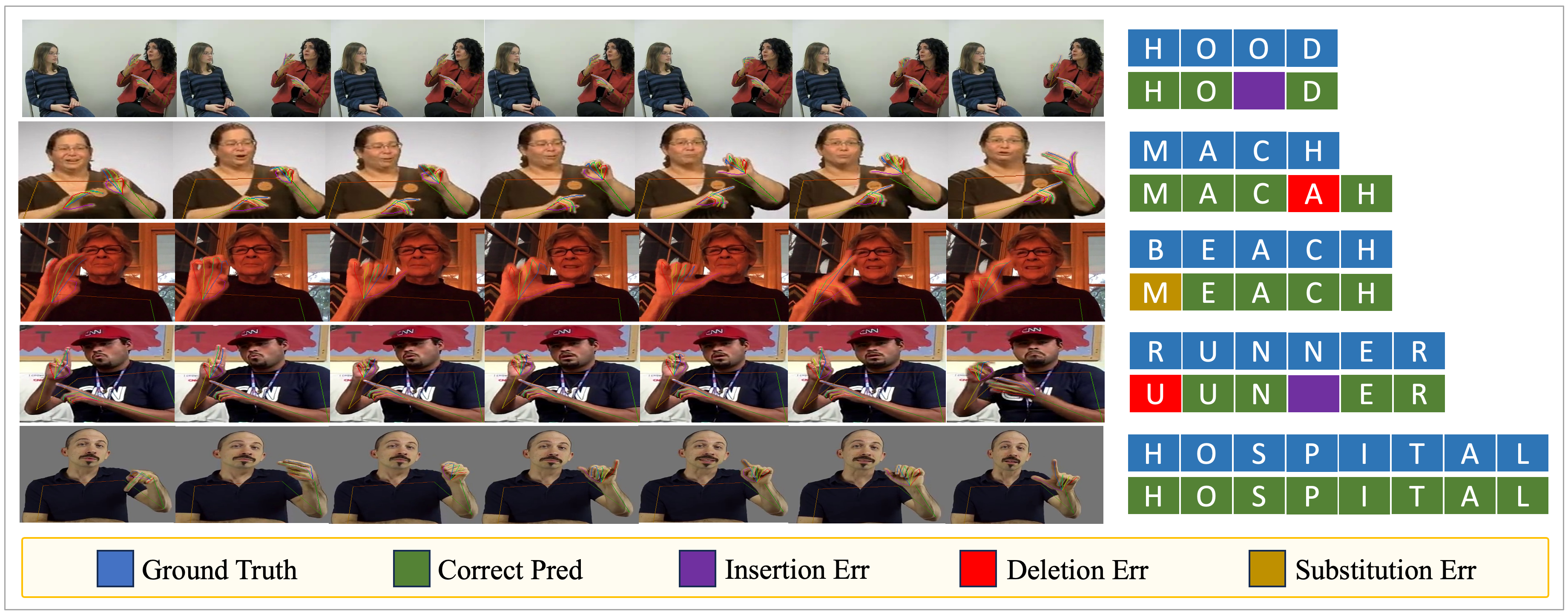}}
\caption{Qualitative results on ChicagoFSWild\cite{fs18slt}. Only a subset of frames is presented here.}
\label{fig:example}
\end{figure*}

\begin{table}[ht]
\centering
\begin{tabular}[t]{lcc}
\toprule
&Letter Accuracy\\
\midrule
OpenPose\cite{openpose}&55.7\\
3D MediaPipe \cite{lugaresi2019mediapipe}&{64.1}\\
Ours without Length Token &{64.1}\\
Ours &\textbf{66.3}\\
\bottomrule
\end{tabular}
\caption{Ablation Analysis of Various Components. Our method is compared with alternative pose estimation approaches. Additionally, we examine the impact of utilizing 3D coordinates instead of 2D, along with assessing the influence of the Length Token.}
\label{tab:Pose}
\end{table}%

\noindent
\subsection{Ablation Study}
In this section, we present a series of ablation studies to evaluate the contribution and effectiveness of various components in our proposed method. Specifically, we investigate the impact of different factors and variations, including the selection of the pose method, diverse decoding formulations used during inference, and the influence of length tokens.
\newline

\noindent
\textbf{Selection of Pose Estimator.} 
\label{ab:Pose}
We begin by analyzing the effect of the Pose Estimator on the overall performance. We evaluate the impact of different pose methods on translation accuracy. We employed OpenPose\cite{openpose} and MediaPipe\cite{lugaresi2019mediapipe} as the pose extractor methods. When comparing MediaPipe\cite{lugaresi2019mediapipe} Holistic to OpenPose\cite{openpose}, notable differences arise in their approach to predicting body keypoints. MediaPipe first predicts the body keypoints and subsequently employs separate models for hand and face keypoints on cropped patches. In contrast, OpenPose predicts all keypoints together from the input image. A distinguishing feature of MediaPipe is its consideration of the consistency between predictions across subsequent frames. This approach promotes smoother and more consistent predictions, reducing the likelihood of detection failures or missed keypoints. Also, MediaPipe directly predicts the keypoints in 3D, offering a more direct estimation. On the other hand, OpenPose relies on triangulation techniques to infer the 3D pose from the detected keypoints. The results of these evaluations are presented in the first row of Table \ref{tab:Pose}. These findings highlight the potential for enhancing the accuracy of the methods by further advancements in pose estimation. Pose models demonstrate greater robustness in handling variations compared to RGB-based methods. Furthermore, pose-based approaches exhibit improved data efficiency during training and also can be advantageous in scenarios where data privacy is a concern.
\\

\noindent
\textbf{3D vs 2D.} 
We further investigate the impact of utilizing 3D coordinates instead of 2D from the Mediapipe\cite{lugaresi2019mediapipe} Holistic approach. The results, presented in the second row of Table \ref{tab:Pose}, indicate a degradation in performance. This suggests that the 3D coordinates may not be reliable and can introduce significant noise to the model.
\\

\noindent
\textbf{Length Token.} 
To assess the impact of the length token in our approach, we conducted experiments where we removed it from the training and decoding process. The third row of Table \ref{tab:Pose} shows the results of the method with/without the Length Token. 


\begin{table}[ht]
\centering
\begin{tabular}[t]{lccc}
\toprule
& Deletions  &  Substitutions & Insertions\\
\midrule
Error Count & 768 & 488 & 231\\
Error Rate & 17.37 & 11.04 & 5.22 \\
\bottomrule
\end{tabular}
\caption{Error Counts and Rates in three Scenarios: Deletions, Substitutions, and Insertions.}
\label{tab:error}
\end{table}%

\begin{figure}[h]
\centerline{\includegraphics[width=0.78\linewidth]{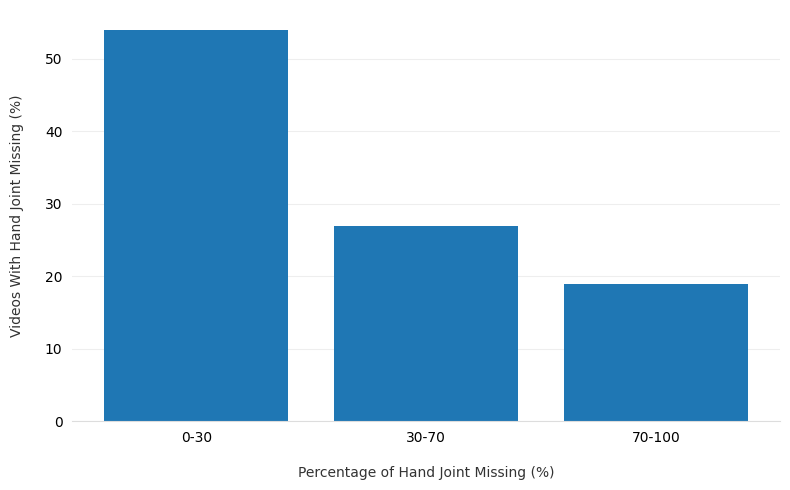}}
\caption{This figure shows the distribution of hand pose availability in the dataset. The x-axis represents the percentage of missing hand poses in each video, while the y-axis indicates the percentage of videos in the dataset falling into each category.}
\label{fig:pose}
\end{figure}

\subsection{Limitations and Failure Cases}
In this section, we discuss the method's failure cases and limitations. The primary errors involve deletions, followed by insertions as shown in Table \ref{tab:error}. Furthermore, regarding substitutions, the top-5 letter pairs that exhibit the highest confusion rates are $(\mathrm{e} \rightarrow \mathrm{o}), (\mathrm{i} \rightarrow \mathrm{y}), (\mathrm{r} \rightarrow \mathrm{u}), (\mathrm{a} \rightarrow \mathrm{o}),(\mathrm{i} \rightarrow \mathrm{j})$. An additional limitation concerns the performance of the pose models employed. Some video frames contain low-quality and frequently blurry images due to fast movements. Figure \ref{fig:pose} shows the distribution of the missing hand joints estimated using openpose\cite{openpose} method. Given that our model relies solely on pose keypoints, instances of failure in the pose model directly lead to the overall failure of our approach. Figure \ref{fig:example} showcases some of the video frames alongside our model's output, displaying both accurate translations and other errors. 

\subsection{Conclusion}
In conclusion, we have presented a novel approach that combines transformer architecture with hand pose models for fingerspelling translation. Our proposed method leverages the language modeling capabilities of transformers while effectively capturing the temporal dynamics of hand poses. Through extensive experiments on the ChicagoWild and ChicagoWild+ datasets, we have demonstrated significant improvements in accuracy and translation performance compared to state-of-the-art models.

\noindent \textbf{Acknowledgments:} This work was supported by the 2023 Amazon Research Awards Program. 

{\small
\bibliographystyle{ieee_fullname}
\bibliography{egbib}

\begin{thebibliography}{10}\itemsep=-1pt

\bibitem{bohavcek2022sign}
Maty{\'a}{\v{s}} Boh{\'a}{\v{c}}ek and Marek Hr{\'u}z.
\newblock Sign pose-based transformer for word-level sign language recognition.
\newblock In {\em Proceedings of the IEEE/CVF Winter Conference on Applications
  of Computer Vision}, pages 182--191, 2022.

\bibitem{camgoz2018neural}
Necati~Cihan Camgoz, Simon Hadfield, Oscar Koller, Hermann Ney, and Richard
  Bowden.
\newblock Neural sign language translation.
\newblock In {\em Proceedings of the IEEE conference on computer vision and
  pattern recognition}, pages 7784--7793, 2018.

\bibitem{camgoz2020sign}
Necati~Cihan Camgoz, Oscar Koller, Simon Hadfield, and Richard Bowden.
\newblock Sign language transformers: Joint end-to-end sign language
  recognition and translation.
\newblock In {\em Proceedings of the IEEE/CVF conference on computer vision and
  pattern recognition}, pages 10023--10033, 2020.

\bibitem{openpose}
Zhe Cao, Tomas Simon, Shih-En Wei, and Yaser Sheikh.
\newblock Realtime multi-person 2d pose estimation using part affinity fields.
\newblock In {\em Proceedings of the IEEE conference on computer vision and
  pattern recognition}, pages 7291--7299, 2017.

\bibitem{subunet}
Necati Cihan~Camgoz, Simon Hadfield, Oscar Koller, and Richard Bowden.
\newblock Subunets: End-to-end hand shape and continuous sign language
  recognition.
\newblock In {\em Proceedings of the IEEE international conference on computer
  vision}, pages 3056--3065, 2017.

\bibitem{cui2017recurrent}
Runpeng Cui, Hu Liu, and Changshui Zhang.
\newblock Recurrent convolutional neural networks for continuous sign language
  recognition by staged optimization.
\newblock In {\em Proceedings of the IEEE conference on computer vision and
  pattern recognition}, pages 7361--7369, 2017.

\bibitem{deng2009imagenet}
Jia Deng, Wei Dong, Richard Socher, Li-Jia Li, Kai Li, and Li Fei-Fei.
\newblock Imagenet: A large-scale hierarchical image database.
\newblock In {\em 2009 IEEE conference on computer vision and pattern
  recognition}, pages 248--255. Ieee, 2009.

\bibitem{du2022full}
Yao Du, Pan Xie, Mingye Wang, Xiaohui Hu, Zheng Zhao, and Jiaqi Liu.
\newblock Full transformer network with masking future for word-level sign
  language recognition.
\newblock {\em Neurocomputing}, 500:115--123, 2022.

\bibitem{duarte2021how2sign}
Amanda Duarte, Shruti Palaskar, Lucas Ventura, Deepti Ghadiyaram, Kenneth
  DeHaan, Florian Metze, Jordi Torres, and Xavier Giro-i Nieto.
\newblock How2sign: a large-scale multimodal dataset for continuous american
  sign language.
\newblock In {\em Proceedings of the IEEE/CVF conference on computer vision and
  pattern recognition}, pages 2735--2744, 2021.

\bibitem{fayyazsanavi2023u2rle}
Pooya Fayyazsanavi, Zhiqiang Wan, Will Hutchcroft, Ivaylo Boyadzhiev, Yuguang
  Li, Jana Kosecka, and Sing~Bing Kang.
\newblock U2rle: Uncertainty-guided 2-stage room layout estimation.
\newblock In {\em Proceedings of the IEEE/CVF Conference on Computer Vision and
  Pattern Recognition}, pages 3561--3569, 2023.

\bibitem{gajurel2021fine}
Kamala Gajurel, Cuncong Zhong, and Guanghui Wang.
\newblock A fine-grained visual attention approach for fingerspelling
  recognition in the wild.
\newblock In {\em 2021 IEEE International Conference on Systems, Man, and
  Cybernetics (SMC)}, pages 3266--3271. IEEE, 2021.

\bibitem{graves2006connectionist}
Alex Graves, Santiago Fern{\'a}ndez, Faustino Gomez, and J{\"u}rgen
  Schmidhuber.
\newblock Connectionist temporal classification: labelling unsegmented sequence
  data with recurrent neural networks.
\newblock In {\em Proceedings of the 23rd international conference on Machine
  learning}, pages 369--376, 2006.

\bibitem{he2019hard}
Tianyu He, Xu Tan, and Tao Qin.
\newblock Hard but robust, easy but sensitive: How encoder and decoder perform
  in neural machine translation.
\newblock {\em arXiv preprint arXiv:1908.06259}, 2019.

\bibitem{hosain2020finehand}
Al~Amin Hosain, Panneer~Selvam Santhalingam, Parth Pathak, Huzefa Rangwala, and
  Jana Kosecka.
\newblock Finehand: Learning hand shapes for american sign language
  recognition, 2020.

\bibitem{hosain2021hand}
Al~Amin Hosain, Panneer~Selvam Santhalingam, Parth Pathak, Huzefa Rangwala, and
  Jana Kosecka.
\newblock Hand pose guided 3d pooling for word-level sign language recognition.
\newblock In {\em Proceedings of the IEEE/CVF Winter Conference on Applications
  of Computer Vision}, pages 3429--3439, 2021.

\bibitem{jiang2021skeleton}
Songyao Jiang, Bin Sun, Lichen Wang, Yue Bai, Kunpeng Li, and Yun Fu.
\newblock Skeleton aware multi-modal sign language recognition.
\newblock In {\em Proceedings of the IEEE/CVF Conference on Computer Vision and
  Pattern Recognition}, pages 3413--3423, 2021.

\bibitem{kabade2023american}
Amruta~E Kabade, Padmashree Desai, C Sujatha, and G Shankar.
\newblock American sign language fingerspelling recognition using attention
  model.
\newblock In {\em 2023 IEEE 8th International Conference for Convergence in
  Technology (I2CT)}, pages 1--6. IEEE, 2023.

\bibitem{kay2017kinetics}
Will Kay, Joao Carreira, Karen Simonyan, Brian Zhang, Chloe Hillier, Sudheendra
  Vijayanarasimhan, Fabio Viola, Tim Green, Trevor Back, Paul Natsev, et~al.
\newblock The kinetics human action video dataset.
\newblock {\em arXiv preprint arXiv:1705.06950}, 2017.

\bibitem{kingma2014adam}
Diederik~P Kingma and Jimmy Ba.
\newblock Adam: A method for stochastic optimization.
\newblock {\em arXiv preprint arXiv:1412.6980}, 2014.

\bibitem{li2020word}
Dongxu Li, Cristian Rodriguez, Xin Yu, and Hongdong Li.
\newblock Word-level deep sign language recognition from video: A new
  large-scale dataset and methods comparison.
\newblock In {\em The IEEE Winter Conference on Applications of Computer
  Vision}, pages 1459--1469, 2020.

\bibitem{li-etal-2023-contrastive-token}
Linjun Li, Tao Jin, Xize Cheng, Ye Wang, Wang Lin, Rongjie Huang, and Zhou
  Zhao.
\newblock Contrastive token-wise meta-learning for unseen performer visual
  temporal-aligned translation.
\newblock In {\em Findings of the Association for Computational Linguistics:
  ACL 2023}, pages 10993--11007, Toronto, Canada, July 2023. Association for
  Computational Linguistics.

\bibitem{li2022multi}
Ronghui Li and Lu Meng.
\newblock Multi-view spatial-temporal network for continuous sign language
  recognition.
\newblock {\em arXiv preprint arXiv:2204.08747}, 2022.

\bibitem{lugaresi2019mediapipe}
Camillo Lugaresi, Jiuqiang Tang, Hadon Nash, Chris McClanahan, Esha Uboweja,
  Michael Hays, Fan Zhang, Chuo-Ling Chang, Ming~Guang Yong, Juhyun Lee, et~al.
\newblock Mediapipe: A framework for building perception pipelines.
\newblock {\em arXiv preprint arXiv:1906.08172}, 2019.

\bibitem{moryossef2021evaluating}
Amit Moryossef, Ioannis Tsochantaridis, Joe Dinn, Necati~Cihan Camgoz, Richard
  Bowden, Tao Jiang, Annette Rios, Mathias Muller, and Sarah Ebling.
\newblock Evaluating the immediate applicability of pose estimation for sign
  language recognition.
\newblock In {\em Proceedings of the IEEE/CVF Conference on Computer Vision and
  Pattern Recognition}, pages 3434--3440, 2021.

\bibitem{nejatishahidin2023graph}
Negar Nejatishahidin, Will Hutchcroft, Manjunath Narayana, Ivaylo Boyadzhiev,
  Yuguang Li, Naji Khosravan, Jana Ko{\v{s}}eck{\'a}, and Sing~Bing Kang.
\newblock Graph-covis: Gnn-based multi-view panorama global pose estimation.
\newblock In {\em Proceedings of the IEEE/CVF Conference on Computer Vision and
  Pattern Recognition}, pages 6458--6467, 2023.

\bibitem{OZ20111204}
Cemil Oz and Ming~C. Leu.
\newblock American sign language word recognition with a sensory glove using
  artificial neural networks.
\newblock {\em Engineering Applications of Artificial Intelligence},
  24(7):1204--1213, 2011.
\newblock Infrastructures and Tools for Multiagent Systems.

\bibitem{padden2003alphabet}
Carol~A Padden and Darline~Clark Gunsauls.
\newblock How the alphabet came to be used in a sign language.
\newblock {\em Sign Language Studies}, pages 10--33, 2003.

\bibitem{pannattee2021novel}
Peerawat Pannattee, Wuttipong Kumwilaisak, Chatchawarn Hansakunbuntheung, and
  Nattanun Thatphithakkul.
\newblock Novel american sign language fingerspelling recognition in the wild
  with weakly supervised learning and feature embedding.
\newblock In {\em 2021 18th International Conference on Electrical
  Engineering/Electronics, Computer, Telecommunications and Information
  Technology (ECTI-CON)}, pages 291--294. IEEE, 2021.

\bibitem{papadimitriou20_interspeech}
Katerina Papadimitriou and Gerasimos Potamianos.
\newblock {Multimodal Sign Language Recognition via Temporal Deformable
  Convolutional Sequence Learning}.
\newblock In {\em Proc. Interspeech 2020}, pages 2752--2756, 2020.

\bibitem{10.1007/978-3-030-66096-3_18}
Maria Parelli, Katerina Papadimitriou, Gerasimos Potamianos, Georgios Pavlakos,
  and Petros Maragos.
\newblock Exploiting 3d hand pose estimation in deep learning-based sign
  language recognition from rgb videos.
\newblock In Adrien Bartoli and Andrea Fusiello, editors, {\em Computer Vision
  -- ECCV 2020 Workshops}, pages 249--263, Cham, 2020. Springer International
  Publishing.

\bibitem{parelli2020exploiting}
Maria Parelli, Katerina Papadimitriou, Gerasimos Potamianos, Georgios Pavlakos,
  and Petros Maragos.
\newblock Exploiting 3d hand pose estimation in deep learning-based sign
  language recognition from rgb videos.
\newblock In {\em Computer Vision--ECCV 2020 Workshops: Glasgow, UK, August
  23--28, 2020, Proceedings, Part II 16}, pages 249--263. Springer, 2020.

\bibitem{paszke2019pytorch}
Adam Paszke, Sam Gross, Francisco Massa, Adam Lerer, James Bradbury, Gregory
  Chanan, Trevor Killeen, Zeming Lin, Natalia Gimelshein, Luca Antiga, et~al.
\newblock Pytorch: An imperative style, high-performance deep learning library.
\newblock {\em Advances in neural information processing systems}, 32, 2019.

\bibitem{prajwal2022weaklyfinger}
KR Prajwal, Hannah Bull, Liliane Momeni, Samuel Albanie, G{\"u}l Varol, and
  Andrew Zisserman.
\newblock Weakly-supervised fingerspelling recognition in british sign language
  videos.
\newblock {\em arXiv preprint arXiv:2211.08954}, 2022.

\bibitem{pu2019iterative}
Junfu Pu, Wengang Zhou, and Houqiang Li.
\newblock Iterative alignment network for continuous sign language recognition.
\newblock In {\em Proceedings of the IEEE/CVF conference on computer vision and
  pattern recognition}, pages 4165--4174, 2019.

\bibitem{ren2020study}
Yi Ren, Jinglin Liu, Xu Tan, Zhou Zhao, Sheng Zhao, and Tie-Yan Liu.
\newblock A study of non-autoregressive model for sequence generation.
\newblock In {\em Proceedings of the 58th Annual Meeting of the Association for
  Computational Linguistics}, pages 149--159, 2020.

\bibitem{santhalingam2019sign}
Panneer~Selvam Santhalingam, Parth Pathak, Jana Ko{\v{s}}eck{\'a}, Huzefa
  Rangwala, et~al.
\newblock Sign language recognition analysis using multimodal data.
\newblock In {\em 2019 IEEE International Conference on Data Science and
  Advanced Analytics (DSAA)}, pages 203--210. IEEE, 2019.

\bibitem{santhalingam2020body}
Panneer~Selvam Santhalingam, Parth Pathak, Jana Ko{\v{s}}eck{\'e}, Huzefa
  Rangwala, et~al.
\newblock Body pose and deep hand-shape feature based american sign language
  recognition.
\newblock In {\em 2020 IEEE 7th International Conference on Data Science and
  Advanced Analytics (DSAA)}, pages 207--215. IEEE, 2020.

\bibitem{schuster1997bidirectional}
Mike Schuster and Kuldip~K Paliwal.
\newblock Bidirectional recurrent neural networks.
\newblock {\em IEEE transactions on Signal Processing}, 45(11):2673--2681,
  1997.

\bibitem{shi2021fingerspelling}
Bowen Shi, Diane Brentari, Greg Shakhnarovich, and Karen Livescu.
\newblock Fingerspelling detection in american sign language.
\newblock In {\em Proceedings of the IEEE/CVF Conference on Computer Vision and
  Pattern Recognition}, pages 4166--4175, 2021.

\bibitem{shi2022open}
Bowen Shi, Diane Brentari, Greg Shakhnarovich, and Karen Livescu.
\newblock Open-domain sign language translation learned from online video.
\newblock In {\em EMNLP}, 2022.

\bibitem{shi-etal-2022-searching}
Bowen Shi, Diane Brentari, Greg Shakhnarovich, and Karen Livescu.
\newblock Searching for fingerspelled content in {A}merican {S}ign {L}anguage.
\newblock In {\em Proceedings of the 60th Annual Meeting of the Association for
  Computational Linguistics (Volume 1: Long Papers)}, pages 1699--1712, Dublin,
  Ireland, May 2022. Association for Computational Linguistics.

\bibitem{fs18slt}
Bowen Shi, Aurora~Martinez Del~Rio, Jonathan Keane, Jonathan Michaux, Diane
  Brentari, Greg Shakhnarovich, and Karen Livescu.
\newblock American sign language fingerspelling recognition in the wild.
\newblock In {\em 2018 IEEE Spoken Language Technology Workshop (SLT)}, pages
  145--152. IEEE, 2018.

\bibitem{fs18iccv}
Bowen Shi, Aurora Martinez~Del Rio, Jonathan Keane, Diane Brentari, Greg
  Shakhnarovich, and Karen Livescu.
\newblock Fingerspelling recognition in the wild with iterative visual
  attention.
\newblock In {\em Proceedings of the IEEE/CVF International Conference on
  Computer Vision}, pages 5400--5409, 2019.

\bibitem{fingeriter}
Bowen Shi, Aurora Martinez~Del Rio, Jonathan Keane, Diane Brentari, Greg
  Shakhnarovich, and Karen Livescu.
\newblock Fingerspelling recognition in the wild with iterative visual
  attention.
\newblock In {\em Proceedings of the IEEE/CVF International Conference on
  Computer Vision}, pages 5400--5409, 2019.

\bibitem{sinha2016deephand}
Ayan Sinha, Chiho Choi, and Karthik Ramani.
\newblock Deephand: Robust hand pose estimation by completing a matrix imputed
  with deep features.
\newblock In {\em Proceedings of the IEEE conference on computer vision and
  pattern recognition}, pages 4150--4158, 2016.

\bibitem{song2021non}
Xingchen Song, Zhiyong Wu, Yiheng Huang, Chao Weng, Dan Su, and Helen Meng.
\newblock Non-autoregressive transformer asr with ctc-enhanced decoder input.
\newblock In {\em ICASSP 2021-2021 IEEE International Conference on Acoustics,
  Speech and Signal Processing (ICASSP)}, pages 5894--5898. IEEE, 2021.

\bibitem{vaswani2017attention}
Ashish Vaswani, Noam Shazeer, Niki Parmar, Jakob Uszkoreit, Llion Jones,
  Aidan~N Gomez, {\L}ukasz Kaiser, and Illia Polosukhin.
\newblock Attention is all you need.
\newblock {\em Advances in neural information processing systems}, 30, 2017.

\bibitem{Wadhawan2021}
Ankita Wadhawan and Parteek Kumar.
\newblock Sign language recognition systems: A decade systematic literature
  review.
\newblock {\em Archives of Computational Methods in Engineering},
  28(3):785--813, May 2021.

\end{thebibliography}
}

\end{document}